\ifdefined\pdfoutput\pdfoutput=1\fi
\documentclass[11pt]{article}

\usepackage[final]{acl}

\usepackage{times}
\usepackage{latexsym}
\usepackage{iftex}
\ifPDFTeX
  \usepackage[T1]{fontenc}
  \usepackage[utf8]{inputenc}
\fi
\usepackage{microtype}
\usepackage{inconsolata}
\usepackage{upquote}   
\usepackage{booktabs}
\usepackage{graphicx}
\usepackage{multirow}
\usepackage{amsmath}
\usepackage{amssymb}
\usepackage{xcolor}
\usepackage{colortbl}
\usepackage{framed}   
\usepackage{enumitem}
\usepackage{url}

\usepackage[textsize=footnotesize]{todonotes}
\newcommand{\code}[1]{\texttt{#1}}
\definecolor{mycolor}{RGB}{0, 110, 180}

\definecolor{deg1}{RGB}{253,237,232}
\definecolor{deg2}{RGB}{249,217,209}
\definecolor{deg3}{RGB}{244,189,178}
\definecolor{deg4}{RGB}{237,151,140}
\definecolor{goodgreen}{RGB}{197,229,197}
\definecolor{gg1}{RGB}{233,245,233}
\definecolor{gg2}{RGB}{213,237,213}
\definecolor{gg3}{RGB}{190,227,190}
\definecolor{gg4}{RGB}{163,213,163}
\definecolor{devblue}{RGB}{214,231,245}
\definecolor{ourgold}{RGB}{250,231,190}
\definecolor{ourgolddark}{RGB}{246,214,146}
\definecolor{selblue}{RGB}{219,233,246}

\definecolor{promptbg}{RGB}{246,246,243}
\definecolor{promptrule}{RGB}{150,164,180}
\newenvironment{promptbox}[1]%
  {\par\addvspace{4pt}%
   \MakeFramed{\advance\hsize-\width \FrameRestore}%
   \setlength{\parindent}{0pt}\setlength{\parskip}{3pt}%
   {\small\textbf{#1}}\par\nobreak\vskip-1pt\nobreak
   \scriptsize\ttfamily\raggedright\relax}%
  {\endMakeFramed\par\addvspace{4pt}}

\ifXeTeX
  \usepackage{fontspec}
  \newfontfamily\arabicfont[Script=Arabic,Scale=1.28,
    Extension=.ttf,UprightFont=*-Regular,BoldFont=*-Bold]{Amiri}
  \newcommand{\ar}[2]{{\arabicfont\beginR #1\endR}}
\else
  \newcommand{\ar}[2]{\textit{#2}}
\fi

\newcommand{\fld}[1]{{\rmfamily\scshape #1}\hspace{0.4em}}

\title{\textcolor{mycolor}{\textsf{\textit{Aslema}}} at NADI 2026: Data Augmentation for Intent Recognition and Slot Filling}

\author{
  Tajwaar Shafiq\thanks{~The contribution was made while the author was
  interning at the Qatar Computing Research Institute.},
  Hunzalah Hassan Bhatti, Firoj Alam, Shammur Absar Chowdhury \\
  Qatar Computing Research Institute, HBKU, Qatar \\
  \texttt{tajwaar.shafiq@alumni.utoronto.ca,} \\
  \texttt{\{hubh90945, fialam, shchowdhury\}@hbku.edu.qa}
}

\begin{document}
\maketitle

\begin{abstract}

We present \textbf{Aslema}, our system for NADI 2026 Shared Task 5, which consists of two subtasks: \textit{intent recognition} and \textit{slot filling}. We evaluate four omni LLMs in a zero-shot setting and compare them with fine-tuned models. Our results show that fine-tuning consistently outperforms zero-shot inference. We further explore synthetic data augmentation by using an LLM to generate culturally grounded Tunisian Derja utterances, followed by voice cloning to generate synthetic speech. Incorporating this synthetic data improves performance on both tasks. Our final submitted system, based on Qwen3-Omni-30B and trained with a mixture of original and synthetic data, achieves 86.8\% intent accuracy and 34.7 WER on the devtest split. On the official test set it ranks \textbf{1st in slot filling} (59.5 CoER) and \textbf{4th among 5 teams in intent recognition} (66.1\% accuracy). We release our experimental scripts\footnote{\url{https://github.com/hunzed/aslema_nadi2026}} and the synthetic dataset\footnote{\url{https://huggingface.co/datasets/QCRI/Aslema-Synth-TN}} to support further research.
\end{abstract}

\section{Introduction}
\label{sec:introduction}

Spoken dialogue interfaces are increasingly driven by large language models (LLMs), with recent \emph{audio} LLMs processing speech directly and combining speech recognition with language understanding~\cite{chu2023qwenaudio,tang2024salmonn,xu2025qwen25omni}. However, their effectiveness for dialectal Arabic remains limited~\cite{abdelali-etal-2024-larabench,al2025landscape,hassan-bhatti-alam-2026-beyond}. Tunisian Dialect (Derja) is low-resource, heavily code-switched with French and English, and substantially different from the Modern Standard Arabic (MSA) that dominates Arabic training corpora~\cite{talafha-etal-2024-casablanca}. 
The NADI shared-task series has increasingly focused on such dialectal settings~\cite{abdul-mageed-etal-2024-nadi,talafha-etal-2025-nadi}. Its 2026 edition~\cite{Sullivan-etal-2026-nadi} introduces Shared Task 5 on end-to-end Spoken Language Understanding (SLU) using SLURP-TN~\cite{elleuch2026slurptn}, a Tunisian re-recording of the SLURP benchmark~\cite{bastianelli-etal-2020-slurp}, covering intent recognition and slot filling.

We participate in both subtasks and study the effectiveness of audio LLMs for dialectal SLU under different levels of supervision. Our experiments cover four instruction-tuned audio LLMs (3B--30B parameters), LoRA-based fine-tuning~\cite{hu2022lora}, and a fully fine-tuned Whisper-small~\cite{radford2023whisper} baseline. We also investigate the effect of training data size and synthetic augmentation using an LLM$+$TTS pipeline.
To summarize, our main contributions are as follows.
\begin{itemize}[noitemsep,topsep=0pt,leftmargin=*,labelsep=.5em]
  \item We provide a systematic evaluation of four audio LLMs for 
  intent recognition and slot filling under zero-shot and LoRA fine-tuning settings.
  \item We develop an LLM$+$TTS augmentation approach that generates culturally grounded Tunisian Derja utterances with slot annotations and converts them to speech using voice-cloned TTS, targeting underrepresented intents.
  \item We evaluate different training configurations using real, synthetic, and mixed real--synthetic data, and use the best-performing configuration to build our submitted system. 
\end{itemize}

\noindent\textbf{Findings.} Our results show that zero-shot audio LLMs perform poorly on both subtasks, while LoRA fine-tuning with $\sim$3 hours of supervised speech substantially improves performance. Synthetic data alone also provides clear gains over zero-shot inference, while combining real and synthetic data achieves the strongest overall results.

\section{Related Work}
\label{sec:related_work}

\paragraph{Spoken language understanding.} SLU maps speech to structured semantics representations. Historically, early SLU systems followed a cascaded design in which an automatic speech recognition (ASR) module produced a transcript that was then processed by a text-based NLU model, in comparison to recent and increasing adoptions to end-to-end architectures~\cite{tur2011slu,ghannay2018e2e-ner,laperriere-etal-2022-media}. This development has been closely accompanied by the release of increasingly challenging benchmarks~\cite{ahmed2026multiturn}. SLURP~\cite{bastianelli-etal-2020-slurp} introduced a single-turn spoken assistant benchmark, while MASSIVE~\cite{fitzgerald-etal-2023-massive} and Speech-MASSIVE~\cite{lee-etal-2024-speech-massive} extended intent and slot annotation to multiple languages. SLURP-TN~\cite{elleuch2026slurptn} further adapts this, re-recording SLURP prompts in Tunisian Derja, complementing existing resources such as TARIC-SLU~\cite{mdhaffar-etal-2024-taric} and TEDxTN~\cite{bougares-etal-2025-tedxtn}.

\noindent\textbf{Audio LLMs.}
Recent audio LLMs such as Qwen-Audio and its Omni successors~\cite{chu2023qwenaudio,xu2025qwen25omni,qwen2025qwen3omni}, SALMONN~\cite{tang2024salmonn}, SpeechGPT~\cite{zhang-etal-2023-speechgpt} and Gemma-4~\cite{gemma2026} combine a speech encoder with an LLM to enable zero-shot SLU via prompting. Although these models achieve strong results on English benchmarks, performance remains noticeably weaker on dialectal and other low-resource speech~\cite{yang-etal-2024-airbench,wang-etal-2025-audiobench,bhatti2026harmonizing,chen2026voicebench}, including Arabic dialects~\cite{talafha-etal-2024-casablanca,alam2025spokennativqa}. Parameter-efficient fine-tuning (PEFT), LoRA in particular~\cite{hu2022lora}, is the standard approach for adapting such models in resource-constrained settings, and previous NADI editions saw adapted speech models consistently outperform approaches in dialectal ASR tracks~\cite{talafha-etal-2025-nadi,munsit-2025-nadi}.

\noindent\textbf{Synthetic data for speech tasks.} Annotated speech data collection remains expensive, particularly for low-resource dialects, pushing for a growing interest in synthetic data generation. LLM-driven data generation in the Self-Instruct paradigm~\cite{wang-etal-2023-self-instruct} has been combined with TTS to construct paired speech-semantic data~\cite{noroozi-etal-2024-instruction}, including recent efforts targeting Arabic~\cite{sheikhali2026menaspeechbank}. We adopt this approach to dialectal SLU, focusing on generating synthetic training samples to increase coverage of underrepresented intent classes in the original training set. On the speech side, we build on VoxCPM~\cite{zhou2026voxcpm2}, a recent open, tokenizer-free TTS model with coverage of 30 languages, including Arabic. Its zero-shot voice cloning capability allows a small number of dialectal reference clips to represent a variety of Arabic dialects, including those without dedicated TTS voices.

\section{Task and Dataset}
\label{sec:dataset}

\subsection{Task Overview}

\textbf{NADI 2026 Shared Task 5} evaluates end-to-end SLU for Tunisian Arabic speech.
The dataset consists of short spoken assistant commands, averaging about four seconds (3.7\,s in training, 4.3-4.5\,s in the evaluation splits).




\textbf{Subtask intent recognition} requires assigning each utterance to a single intent label. The released training data contains 23 intent labels, while the blind test set follows up to the full 60-label SLURP intent label set. The primary evaluation metrics are accuracy and weighted-F$_1$, as used by the official Codabench leaderboard. We additionally report macro-F$_1$ to better capture performance on underrepresented intent classes. These metrics provide complementary views: accuracy and weighted-F$_1$ are influenced by frequent classes, whereas macro-F$_1$ gives equal weight to each intent class.

\textbf{Subtask slot filling} requires a transcription with inline slot annotations (\code{<label> value >}). The primary evaluation metrics are concept error rate (CoER) and concept-value error rate (CVER), which measure errors in slot labels and their associated values. We additionally report word error rate (WER) and character error rate (CER) on the lexical content after removing the slot markup. These metrics capture complementary aspects of performance: a system may transcribe the Derja utterance correctly while assigning incorrect slot labels or boundaries. Reporting both distinguishes transcription from semantic annotation errors. All scores use the organizers' SLURP-TN baseline evaluation toolkit~\cite{elleuch2026slurptn}.




\subsection{Dataset}
\label{sec:data}

\subsubsection{SLURP-TN Dataset.} SLURP-TN~\cite{elleuch2026slurptn} contains Tunisian Derja re-recordings of SLURP assistant commands, with intent and slot annotations transferred from the original dataset. In Table~\ref{tab:data}, we summarize the data splits. The dataset is relatively small, with $\sim$2.8 hours of training speech, and exhibits substantial class imbalance. The released training set contains 23 intent labels, although only 21 are observed in the training split. Among these, six intents have fewer than 10 training examples, and three are absent from the devtest split. In addition, the most frequent intent accounts for 18.8\% of the devtest utterances.

\begin{table}[t]
\centering
\small
\setlength{\tabcolsep}{4pt}
\begin{tabular}{lrrrr}
\toprule
\textbf{Split} & \textbf{Utts.} & \textbf{Hours} & \textbf{Intents} & \textbf{Slot} \\
\midrule
Train & 2,677 & 2.78 & 21 & 66.2\% \\
Dev   &   595 & 0.74 & 19 & 65.4\% \\
Devtest &   893 & 1.06 & 18 & 62.4\% \\
Test  &   989 & 1.17 & -- & -- \\
\bottomrule
\end{tabular}
\vspace{-0.3cm}
\caption{Dataset statistics. \emph{Intents}: distinct labels present;
\emph{Slotted}: utterances with a gold slot. \emph{Test} gold labels are
not public.}
\label{tab:data}
\vspace{-0.3cm}
\end{table}



\subsubsection{Data augmentation.}
\begin{figure}[t]
\centering
{\includegraphics[width=0.95\columnwidth]{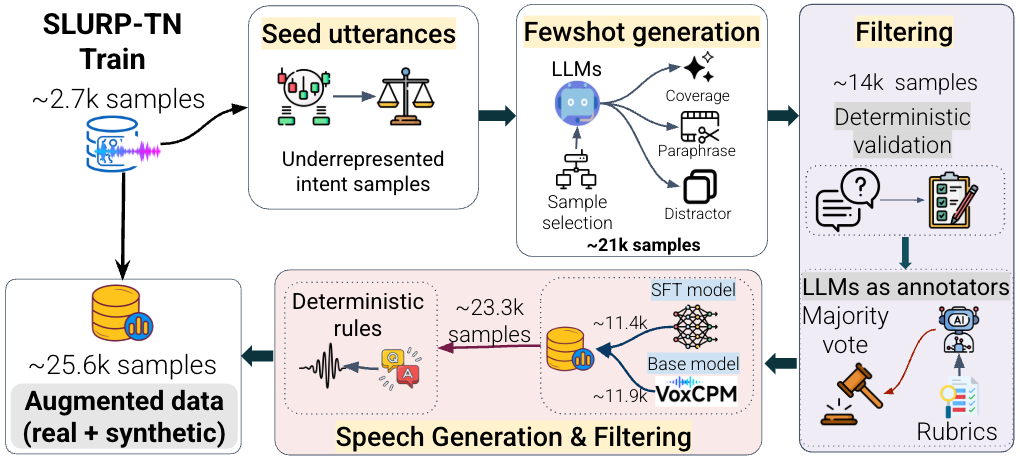}}
\vspace{-0.2cm}
\caption{Overview of the data augmentation pipeline.
}
\label{fig:pipeline}
\vspace{-0.2cm}
\end{figure}

In Figure~\ref{fig:pipeline}, we show an overview of our data augmentation pipeline. We augment the training set to increase the number of examples for underrepresented intents among the 23 training labels. 

\noindent\textbf{Seed utterances.} We first determine the number of synthetic examples for each intent based on its frequency in the training set, generating more examples for intents with fewer training instances. 

\noindent\textbf{Fewshot generation.} We use Gemini 3.1 Pro for the most underrepresented intents and Gemini 3.6 Flash for the remaining intents. Following Self-Instruct~\cite{wang-etal-2023-self-instruct,noroozi-etal-2024-instruction}, the models generate Tunisian Derja utterances with inline slot annotations using six few-shot examples drawn exclusively from the training split. We generate examples in three ways: \textit{(i)} creating new utterances for an intent, \textit{(ii)} paraphrasing existing examples, and \textit{(iii)} generating more challenging examples with the same intent and no slot values. They contribute 16,387, 1,510, and 3,040 of the 20,937 raw candidates respectively 

\noindent\textbf{Filtering.} Since generated utterances may contain duplicates, formatting errors, or dialectal inconsistencies, we apply rule-based approach to identify and remove such cases. This step reduces the set to \textbf{13,876} utterances. 
Finally, three LLMs independently evaluate each utterance for Derja naturalness, intent consistency, and slot correctness using the same evaluation criteria. We keep samples accepted by at least two of the three models, resulting in \textbf{12,138} synthetic utterances. Appendix~\ref{app:augpipeline} provides further details on the generation strategy, validation checks, and LLMs used at each stage (Table~\ref{tab:llms}); the corresponding generation and judging prompts are released with our experimental scripts.

\noindent\textbf{Speech generation and filtering.} We synthesize speech for each utterance using VoxCPM~\cite{zhou2026voxcpm2} under two settings: the original model and a VoxCPM model fine-tuned with LoRA on the $\sim$2.8 hours of SLURP-TN training speech. For voice cloning, both models use the same pool of 152 reference utterances with zero WER, 
selected by an LLM-based ASR check from all 2,330 training clips of 2.5-10s. We then filter the generated speech using rule-based criteria for duration, signal level, clipping, voiced-frame activity, and speaking rate. We generate \textbf{23,300} total utterances (11,946 from the base model and 11,354 from the LoRA-fine-tuned model), of which \textbf{22,940} retained after filtering. Mixing these with original 2,677 results in an augmented training set of \textbf{25,617} utterances.


%

\section{System}
\label{sec:system}

\paragraph{Models.}
We evaluate four instruction-tuned audio LLMs:
\code{Qwen2.5-Omni-3B}, \code{Qwen2.5-Omni-7B}~\cite{xu2025qwen25omni},
\code{Qwen3-Omni-30B-A3B-Instruct}~\cite{qwen2025qwen3omni}, and
\code{gemma-4-E4B-it}~\cite{gemma2026}, served through
ms-swift~\cite{zhao2025msswift} with a vLLM backend~\cite{kwon2023vllm} on a
single H200 GPU, with greedy decoding. We additionally train
\code{whisper-small}~\cite{radford2023whisper} as a small-model baseline.


\noindent
\textbf{LoRA fine-tuning.}
We fine-tune each model on the SLURP-TN training split using LoRA~\cite{hu2022lora}, while keeping the audio encoder and audio--text aligner frozen. We use the same prompts and output formats as in the zero-shot setting to ensure a direct comparison. For each model, we train a single LoRA adapter jointly on data from both subtasks and evaluate it separately on intent recognition and slot filling. We train all models for two epochs and use the final checkpoint, without selecting checkpoints based on held-out loss. Appendix~\ref{app:hyperparams} provides the full hyperparameters, model merging and inference setup, and the Whisper baseline, while Appendix~\ref{app:prompts} provides the subtask prompts.

\noindent
\textbf{Final system.}
We select Qwen3-Omni-30B-A3B for our submitted system, and fine-tune it for two epochs on the combined real and synthetic data.

\section{Results}
\label{sec:results}

In Table~\ref{tab:main}, we report results on the dev-test and official test sets. We provide additional model-level results in Table ~\ref{tab:grid}, and error analyses in Appendix~\ref{app:diagnostics}.

\noindent
\textbf{Official test results.}
Our final system ranked \textbf{1st} in slot filling (Subtask 5.2) with a CoER
of 59.5 and a CVER of 94.2. On intent recognition (Subtask 5.1) the same system initially scored 30.4\% accuracy,
against 86.8\% on dev-test. This large gap is what led us to
inspecting the prediction distribution rather than the model.

We found the SLURP-TN training data covers six scenarios (alarm, email, general, news,
takeaway, weather), and a 23-label
intent inventory, while the official test set is separately drawn from the full SLURP
taxonomy of up to 60 labels over 18 scenarios. This mismatch went unnoticed by us until the blind-test submission. Table~\ref{tab:quirky} contrasts the system's predictions on dev-test and
blind test sets.

For an unseen intent, the task rules score both an abstention (predicted as \code{unknown}) or an exact label as correct. To make this abstention explicit, we deterministically keep a prediction unchanged when it falls into a trained scenarios, and map every other prediction, including \code{general\_quirky}, to \code{unknown}. This simple rule (Appendix~\ref{app:open}) raises accuracy by 35.7 points to \textbf{66.1\%} (66.9 weighted-F$_1$),
ranking \textbf{4th of 5 teams}.

\begin{table}[t]
\centering
\small
\setlength{\tabcolsep}{6pt}
\scalebox{0.99}{
\begin{tabular}{lcc}
\toprule
\textbf{Model} & \textbf{Intent} $\uparrow$ & \textbf{Slot} $\downarrow$ \\
\midrule
\multicolumn{3}{c}{\textbf{Dev-test} (23 labels)} \\   \midrule
Qwen2.5-Omni-3B & 29.2 & 125.2 \\
Qwen2.5-Omni-7B & 42.3 & 150.1 \\
Qwen3-Omni-30B & 52.5 & 131.0 \\
Gemma-4-E4B-it & 53.1 & 97.5 \\
Whisper-small FT & 67.4 & 81.4 \\
\rowcolor{devblue}Qwen2.5-Omni-3B FT & 80.5 & 57.0 \\
\rowcolor{devblue}Qwen2.5-Omni-7B FT & 81.3 & 51.9 \\
\rowcolor{devblue}Gemma-4-E4B-it FT & 80.4 & 49.1 \\
\rowcolor{devblue}Qwen3-Omni-30B FT & 82.9 & 47.7 \\
\rowcolor{ourgolddark}Qwen3-Omni-30B FT (\textit{Mix}) & \textbf{86.8} & \textbf{36.9} \\
\rowcolor{ourgolddark}Qwen3-Omni-30B FT (\textit{Synth}) & 75.6 & 62.2 \\
\midrule
\rowcolor{goodgreen}\multicolumn{3}{c}{\textbf{Blind test set} (60 labels)} \\ \midrule
\rowcolor{goodgreen}Our system (\textit{Mix}) & \underline{\textbf{66.1}} & \underline{\textbf{59.5}} \\
\bottomrule
\end{tabular}
}
\vspace{-0.2cm}
\caption{Results for both subtasks across different splits. Intent recognition is evaluated using accuracy (higher is better), while slot filling is evaluated using concept error rate (lower is better). FT: fine-tuned.}
\label{tab:main}
\vspace{-0.2cm}
\end{table}


\noindent
\textbf{Zero-shot vs. baseline models.}
As shown in Table~\ref{tab:main}, the four omni models show limited zero-shot performance, reaching 29.2--53.1\% intent accuracy and 97.5--150.1 CoER. In comparison, Whisper-small, fully fine-tuned on the training set, achieves 67.4\% intent accuracy and 81.4 CoER, outperforming all zero-shot omni models on both subtasks. This result shows that task-specific fine-tuning with fewer than three hours of Tunisian Derja speech is more effective than direct zero-shot inference with recent omni models. In Table~\ref{tab:grid}, we provide additional analysis of their zero-shot behavior.

\noindent
\textbf{Effect of LoRA fine-tuning.}
We fine-tune each omni model with LoRA on the training split for both subtasks. As shown in Table~\ref{tab:main}, fine-tuning consistently improves all four models. Intent accuracy increases from 29.2-53.1\% to 80.4-82.9\%, while slot CoER decreases from 97.5-150.1 to 47.7-57.0. Qwen3-Omni-30B achieves the best performance after fine-tuning, with 82.9\% intent accuracy and 47.7 CoER. Additionally, fine-tuning reduces the performance gap across model sizes. The spread in intent accuracy among the 3B--30B models decreases from 23.9 points in the zero-shot setting to only 2.5 points after fine-tuning. This suggests that task-specific adaptation substantially reduces the advantage of larger models on the in-domain dev-test set.

\noindent
\textbf{Effect of data augmentation.}
We further fine-tune the best-performing model, Qwen3-Omni-30B-A3B, using two data settings. \emph{Mix} combines 2,677 real and 22,940 synthetic utterances, while \emph{Synth} uses only synthetic speech. As shown in Table~\ref{tab:main}, \emph{Mix} improves both subtasks over fine-tuning on real speech, increasing intent accuracy by 3.9 points and reducing CoER by 10.8 points. \emph{Synth} performs better than zero-shot inference but remains below fine-tuning on real speech. These results show that synthetic speech is more effective when combined with real data than when used alone.

\section{Conclusions and Future Work}
\label{sec:conclusions}

We participated in both subtasks of NADI 2026 Shared Task 5. Our experiments show that current audio LLMs have limited zero-shot performance on Tunisian Derja SLU, and increasing model scale alone does not overcome this limitation. Fine-tuning on fewer than three hours of real speech improves performance to 82.9\% intent accuracy and 40.0 WER for slot filling. Augmenting the training data with synthetic speech further improves performance to 86.8\% intent accuracy and 34.7 WER. Our final system ranks \textbf{1st} in slot filling and \textbf{4th} of five teams in intent recognition. Future work will extend the pipeline to other Arabic dialects, incorporate human validation into synthetic-data filtering, and release a human-validated subset for further analysis and auditing.

\section*{Limitations}
Our evaluation is primarily based on the dev-test split, with the final setup evaluated only on the official blind test set. In addition, the released splits cover fewer intent labels than the official test set, making generalization challenging. Our blind-test intent score also depends on an abstention rule (Appendix~\ref{app:open}) designed after observing the initial performance gap relative to dev testing. The real-only and synthetic configurations are also trained for the same number of epochs on different-sized training sets, so the mixed run trains on roughly 10 times as many optimisation steps, and its gain cannot be attributed to the synthetic data alone. Quality control on the generated data also rests on LLM judges rather than native speakers, so a small-scale native-speaker evaluation remains necessary. Finally, we evaluate synthetic augmentation only with Qwen3-Omni-30B-A3B due to computational constraints.  Extending this analysis to smaller models and incorporating human validation of synthetic speech are promising directions for future work.

\bibliography{refs}

\appendix

\section{Additional Results}
\label{sec_app_additional_results}

In Table~\ref{tab:grid}, we report detailed results for intent recognition and slot filling under both zero-shot and fine-tuned settings. In Table~\ref{tab:synth}, we further analyze the effect of synthetic data and training duration on Qwen3-Omni-30B-A3B. Combining real and synthetic speech performs better than using either source alone, while increasing training from 2 to 2.5 epochs provides only marginal gains.

\begin{table}[!tbh]
\centering
\scriptsize
\setlength{\tabcolsep}{4pt}

\begin{tabular}{lccc}
\toprule
\multicolumn{4}{c}{\textbf{Intent Recognition}} \\
\midrule
\textbf{Model} & \textbf{Acc} $\uparrow$
& \textbf{M-F$_1$} $\uparrow$ & \textbf{W-F$_1$} $\uparrow$ \\
\midrule
\multicolumn{4}{c}{\textbf{Dev-test} (23 labels)} \\ \midrule
Qwen2.5-Omni-3B & 29.2 & 22.6 & 34.4 \\
Qwen2.5-Omni-7B & 42.3 & 29.3 & 45.3 \\
Qwen3-Omni-30B & 52.5 & 38.9 & 57.6 \\
Gemma-4-E4B-it & 53.1 & 46.0 & 59.2 \\
Whisper-small FT & 67.4 & 47.7 & 70.4 \\
\rowcolor{devblue}Qwen2.5-Omni-3B FT & 80.5 & 54.9 & 79.7 \\
\rowcolor{devblue}Qwen2.5-Omni-7B FT & 81.3 & 57.0 & 80.6 \\
\rowcolor{devblue}Gemma-4-E4B-it FT & 80.4 & 56.5 & 79.8 \\
\rowcolor{devblue}Qwen3-Omni-30B FT & 82.9 & 58.0 & 82.4 \\
\rowcolor{ourgolddark}Qwen3-Omni-30B FT (\textit{Mix}) & \textbf{86.8} & \textbf{61.9} & \textbf{86.3} \\
\rowcolor{ourgolddark}Qwen3-Omni-30B FT (\textit{Synth}) & 75.6 & 56.9 & 79.0 \\
\midrule
\rowcolor{goodgreen}\multicolumn{4}{c}{\textbf{Blind test set} (60 labels)} \\ \midrule
\rowcolor{goodgreen}Our system (\textit{Mix}) & \underline{\textbf{66.1}} & -- & \underline{\textbf{66.9}} \\
\bottomrule
\end{tabular}

\vspace{0.15cm}

\begin{tabular}{lcccc}
\toprule
\multicolumn{5}{c}{\textbf{Slot Filling}} \\
\midrule
\textbf{Model} & \textbf{WER} $\downarrow$
& \textbf{CER} $\downarrow$ & \textbf{CoER} $\downarrow$
& \textbf{CVER} $\downarrow$ \\
\midrule
\multicolumn{5}{c}{\textbf{Dev-test} (23 labels)} \\ \midrule
Qwen2.5-Omni-3B & 121.9 & 120.1 & 125.2 & 134.7 \\
Qwen2.5-Omni-7B & 104.7 & 102.5 & 150.1 & 159.6 \\
Qwen3-Omni-30B & 87.8 & 79.6 & 131.0 & 144.7 \\
Gemma-4-E4B-it & 68.3 & 40.6 & 97.5 & 102.1 \\
Whisper-small FT & 56.2 & 25.1 & 81.4 & 107.2 \\
\rowcolor{devblue}Qwen2.5-Omni-3B FT & 47.3 & 18.9 & 57.0 & 97.5 \\
\rowcolor{devblue}Qwen2.5-Omni-7B FT & 45.3 & 18.0 & 51.9 & 91.6 \\
\rowcolor{devblue}Gemma-4-E4B-it FT & 41.4 & 16.4 & 49.1 & 89.1 \\
\rowcolor{devblue}Qwen3-Omni-30B FT & 40.0 & 15.6 & 47.7 & 84.7 \\
\rowcolor{ourgolddark}Qwen3-Omni-30B FT (\textit{Mix}) & \textbf{34.7} & \textbf{12.5} & \textbf{36.9} & \textbf{73.6} \\
\rowcolor{ourgolddark}Qwen3-Omni-30B FT (\textit{Synth}) & 60.6 & 35.0 & 62.2 & 103.2 \\
\midrule
\rowcolor{goodgreen}\multicolumn{5}{c}{\textbf{Blind test set} (60 labels)} \\ \midrule
\rowcolor{goodgreen}Our system (\textit{Mix}) & -- & -- & \underline{\textbf{59.5}} & \underline{\textbf{94.2}} \\
\bottomrule
\end{tabular}

\caption{Detailed results for both subtasks across different splits. Intent recognition is evaluated using accuracy, macro-F$_1$ and weighted-F$_1$ (higher is better), while slot filling is evaluated using WER, CER, CoER and CVER (lower is better). \emph{Mix} combines real and synthetic training data, while \emph{Synth} uses synthetic data only. FT: fine-tuned.}
\label{tab:grid}
\end{table}

\begin{table}[!tbh]
\centering
\small
\setlength{\tabcolsep}{3.2pt}
\begin{tabular}{llccccc}
\toprule
\textbf{Train data} & \textbf{Ep.} & \textbf{Acc} & \textbf{WER} & \textbf{CER} & \textbf{CoER} & \textbf{CVER} \\
\midrule
\multicolumn{7}{l}{\emph{Dev-test split}} \\
Real only  & 2   & \cellcolor{deg1}82.9 & \cellcolor{deg2}40.0 & \cellcolor{deg3}15.6 & \cellcolor{deg3}47.7 & \cellcolor{deg2}84.7 \\
\rowcolor{selblue}
Mix        & 2   & 86.8 & \textbf{34.7} & 12.5 & 36.9 & \textbf{73.6} \\
Mix        & 2.5 & \textbf{87.2} & 34.8 & \textbf{12.3} & \textbf{36.2} & 75.0 \\
Synth only & 2.5 & \cellcolor{deg2}75.6 & \cellcolor{deg4}60.6 & \cellcolor{deg4}35.0 & \cellcolor{deg4}62.2 & \cellcolor{deg3}103.2 \\
\midrule
\multicolumn{7}{l}{\emph{Validation split}} \\
\rowcolor{selblue}
Mix        & 2   & \textbf{86.1} & 32.0 & \textbf{10.8} & \textbf{34.2} & \textbf{66.0} \\
Mix        & 2.5 & 85.7 & \textbf{31.6} & \textbf{10.8} & 35.5 & 67.8 \\
\bottomrule
\end{tabular}
\caption{Effect of synthetic augmentation on Qwen3-Omni-30B-A3B. \emph{Mix} combines real and synthetic training data, while \emph{Synth} uses synthetic data only.}
\label{tab:synth}
\end{table}


\newpage

\section{Error Analysis}
\label{app:diagnostics}

\paragraph{Zero-shot slot-filling errors.}
Zero-shot models often fail to follow the required slot markup format. Fewer than 11\% of their outputs use the expected \code{<label> value >} structure, with missing markup as the dominant error. Models also occasionally generate alternative XML-style forms or over-generate slot content. After fine-tuning, however, the markup rate closely matches the gold distribution, showing that task-specific adaptation largely resolves these formatting errors.

\paragraph{Effect of intent-label granularity.}
Some zero-shot intent errors reflect confusion among semantically related labels rather than complete misunderstanding of the utterance. Mapping intents to coarser action labels improves zero-shot accuracy by about 7-8 points for the strongest models, while the same mapping yields only a small gain after fine-tuning. This suggests that fine-tuning helps models distinguish closely related intent labels more reliably.

\paragraph{Effect of data augmentation.}
We also examine how synthetic augmentation changes errors on the dev-test set. Fine-tuning on real speech corrects a large portion of the zero-shot errors, and adding synthetic speech provides further gains on both subtasks. Overall, augmentation improves intent accuracy by 3.9 points and reduces CoER by 10.8 points (Table~\ref{tab:synth}).

The gains are strongest for less frequent intents and slot types. Common intents show only small improvements, whereas rarer intents gain substantially more (Table~\ref{tab:augdelta}). We observe the same trend for slot filling, where utterances containing less frequent slot types benefit considerably more from augmentation than those containing only common slot types.

\section{Data Augmentation Pipeline}
\label{app:augpipeline}

\paragraph{Deterministic filtering.}
We first apply deterministic filters to the generated utterances. We retain candidates that follow the annotation format, align with the annotated text, use valid task labels, contain sufficient Arabic-script content, and fall within a 2-25 token range. We also remove character 3-gram near-duplicates of real training utterances and previously accepted synthetic examples. These filters reduce malformed, out-of-domain, and repetitive generations before LLM-based evaluation.

\paragraph{Examples from the filtering pipeline.}
Figure~\ref{fig:pipeline} summarizes the complete augmentation process. Below, we illustrate three representative outcomes: rejection during text validation, rejection after speech synthesis, and acceptance into the final training set.

\begin{promptbox}{(a) Rejected during text validation}
\fld{intent} email\_addcontact

\fld{text} \ar{نحب نزيد مهدي لل carnet d adresses}{n\d{h}ebb nzid Mehdi lel carnet d adresses}

\fld{gloss} ``I want to add Mehdi to the address book.''

\rmfamily\upshape
\textbf{Rejected:} the Arabic-script ratio falls below the required threshold because most of the utterance is written in French.
\end{promptbox}

\begin{promptbox}{(b) Rejected after speech synthesis}
\fld{intent} general\_greet

\fld{text} \ar{جمعة مبروكة}{jom\{a mabruka}

\fld{gloss} ``Blessed Friday.''

\rmfamily\upshape
\textbf{Text accepted; audio rejected:} the generated speech is unusually long for the short utterance and fails the speaking-rate filter.
\end{promptbox}

\begin{promptbox}{(c) Accepted into the training set}
\fld{intent} alarm\_query

\fld{text} \ar{فماشي alarme مبرمجة متاع القايلة}{famma\v{s}i alarme mbarmja mt\={a}\{ el-q\={a}yla}

\fld{gloss} ``Is there an alarm set for the early afternoon?''

\rmfamily\upshape
\textbf{Accepted:} the text passes all validation and judging stages, and both synthesized renditions satisfy the acoustic filters.
\end{promptbox}

\paragraph{LLM roles.}
We use different LLMs for complementary stages of the pipeline, as summarized in Table~\ref{tab:llms}. Gemini 3.6 Flash generates the bulk of the synthetic data, while Gemini 3.1 Pro focuses on lower-resource intents and provides reference ASR checks. We use all three models as independent judges and retain an utterance when at least two judges accept it. This majority-vote stage retains 12,138 of the 13,876 candidates that reach LLM-based validation.

\begin{table}[!tbh]
\centering
\small
\setlength{\tabcolsep}{3.5pt}
\scalebox{0.95}{
\begin{tabular}{lcccc}
\toprule
& \multicolumn{2}{c}{\textbf{Generation}} & \textbf{Judging} & \textbf{Ref.} \\
\cmidrule(lr){2-3}
\textbf{Model} & \textbf{Rare} & \textbf{Bulk} & \textbf{panel} & \textbf{ASR} \\
\midrule
Gemini 3.6 Flash & $\times$ & \checkmark & \checkmark & $\times$ \\
\quad\emph{stage output} & -- & 16,637 & 85.4\% & -- \\
Gemini 3.1 Pro   & \checkmark & $\times$ & \checkmark & \checkmark \\
\quad\emph{stage output} & 4,300 & -- & 77.6\% & 2,330 \\
Gemini 2.5 Pro   & $\times$ & $\times$ & \checkmark & $\times$ \\
\quad\emph{stage output} & -- & -- & 74.6\% & -- \\
\bottomrule
\end{tabular}
}
\vspace{-0.2cm}
\caption{LLM roles in the augmentation pipeline. The judging percentages indicate the share of candidates accepted by each model. A 2-of-3 majority vote retains 12,138 of the 13,876 judged candidates.}
\label{tab:llms}
\vspace{-0.3cm}
\end{table}

\paragraph{Effect on label coverage.}
Synthetic augmentation also makes the training distribution more balanced across intents. As shown in Table~\ref{tab:augdelta}, the largest F$_1$ gains occur for less frequent labels, while already frequent intents such as \code{weather\_query} and \code{news\_query} change only slightly. Across the 13 intent labels with sufficient dev-test instances, macro-F$_1$ improves from 80.3 to 85.8. This pattern suggests that augmentation primarily improves coverage of underrepresented intents rather than further emphasizing already common classes.

\begin{table}[t]
\centering
\small
\setlength{\tabcolsep}{2.8pt}
\scalebox{0.95}{
\begin{tabular}{lr|rr|rrr}
\toprule
& & \multicolumn{2}{c|}{\textbf{Train \%}} & \multicolumn{3}{c}{\textbf{F$_1$}} \\
\cmidrule(lr){3-4}\cmidrule(lr){5-7}
\textbf{Intent label} & \textbf{Dev.} & \textbf{Real} & \textbf{Mix} & \textbf{Real} & \textbf{Mix} & \multicolumn{1}{c}{\textbf{$\Delta$}} \\
\midrule
general\_quirky     & 168 & \cellcolor{deg3}15.9 & 6.2 & 75.1 & 80.2 & \cellcolor{gg3}$+5.2$ \\
weather\_query      & 152 & \cellcolor{deg4}18.8 & 6.0 & 89.0 & 89.6 & \cellcolor{gg1}$+0.6$ \\
news\_query         & 122 & \cellcolor{deg4}18.7 & 6.2 & 88.7 & 89.2 & \cellcolor{gg1}$+0.4$ \\
email\_query        & 117 & \cellcolor{deg2}8.1 & 5.4 & 89.9 & 95.7 & \cellcolor{gg3}$+5.7$ \\
email\_sendemail    & 113 & \cellcolor{deg2}6.7 & 4.8 & 88.0 & 92.7 & \cellcolor{gg2}$+4.8$ \\
alarm\_set          & 41  & \cellcolor{deg2}6.8 & 5.3 & 86.4 & 90.2 & \cellcolor{gg2}$+3.9$ \\
alarm\_query        & 34  & \cellcolor{deg1}4.9 & 8.2 & 83.1 & 90.9 & \cellcolor{gg4}$+7.8$ \\
takeaway\_query     & 32  & \cellcolor{deg1}4.5 & 7.1 & 80.6 & 84.4 & \cellcolor{gg2}$+3.8$ \\
email\_querycontact & 26  & 2.4 & 5.7 & 80.8 & 88.0 & \cellcolor{gg4}$+7.2$ \\
takeaway\_order     & 22  & \cellcolor{deg1}5.0 & 6.5 & 72.7 & 80.0 & \cellcolor{gg4}$+7.3$ \\
alarm\_remove       & 21  & 2.9 & 7.4 & 75.0 & 83.7 & \cellcolor{gg4}$+8.7$ \\
general\_joke       & 16  & 2.0 & 7.9 & 71.0 & 78.8 & \cellcolor{gg4}$+7.8$ \\
email\_addcontact   & 12  & 1.1 & \cellcolor{goodgreen}8.2 & 64.0 & 71.4 & \cellcolor{gg4}$+7.4$ \\
\midrule
\emph{Macro, 13 labels} & & & & 80.3 & \textbf{85.8} & \cellcolor{gg3}$\mathbf{+5.4}$ \\
\bottomrule
\end{tabular}
}
\vspace{-0.2cm}
\caption{Effect of augmentation across intent labels, sorted by dev-test support. \emph{Dev.} denotes the number of gold dev-test utterances for each label.}
\label{tab:augdelta}
\vspace{-0.2cm}
\end{table}

\section{Hyperparameters}
\label{app:hyperparams}

We use LoRA with rank 16 and $\alpha=32$, applied only to the attention projections of the language backbone. We set the learning rate to $10^{-4}$ and the effective batch size to 8, while keeping the audio encoder and audio-text aligner frozen. After fine-tuning, we merge the LoRA adapter into the base model and use greedy decoding with the same vLLM serving setup. We fully fine-tune Whisper-small for two epochs and prepend a \code{[INTENT]} or \code{[SLOT]} task marker during decoding. We train all fine-tuned systems for two epochs, as extending mixed-data training to 2.5 epochs provides no consistent improvement (Table~\ref{tab:synth}).

\section{Label-Space Domain Expansion on the Blind Test}
\label{app:open}

\begin{table}[t]
\centering
\small
\begin{tabular}{lcc}
\toprule
 & \textbf{Dev-test} & \textbf{Blind test} \\
 & (893) & (989) \\
\midrule
Gold \code{general\_quirky}      & 18.8\% & n/a \\
Predicted \code{general\_quirky} & 20.8\% & 56.8\% \\
Predictions within 23 labels     & 100\%  & 99.7\% \\
\midrule
Accuracy (as submitted)          & 86.8\% & 30.4\% \\
Accuracy (with abstention)       & --     & 66.1\% \\
\bottomrule
\end{tabular}
\caption{Prediction-distribution for the submitted system
(Qwen3-Omni-30B-A3B, \emph{Mix}, 2-epochs).}
\label{tab:quirky}
\end{table}

\paragraph{The rule.}
For each utterance we take the system's own predicted intent and apply, in order:

\begin{enumerate}\itemsep0pt
\item If the prediction names one of the six trained scenarios and is not
\code{general\_quirky}, keep it unchanged (424 utterances).
\item Otherwise, including \code{general\_quirky} and any label outside the six
scenarios, emit \code{unknown} (565 utterances).
\end{enumerate}

\noindent
The invariant is that no prediction is ever replaced by a different intent: every
change is a label becoming an abstention. The rule has no tunable threshold, uses
no test-set labels, and is applied to the unchanged predictions of the submitted
checkpoint.

\paragraph{Limitations of the rule.}
We cannot decompose the 66.1\% into correct abstentions and correct in-domain
labels, because the test gold annotation is not released. The 35.7-point gain
consists entirely of labels becoming abstentions, so we predict at least 353 of 989
utterances (35.7\%) carry a gold intent outside the released inventory; the true
figure is higher by however many abstentions were themselves in-domain errors.

\section{Prompts}
\label{app:prompts}

We use separate prompts for intent recognition and slot filling. For the official intent-recognition test set, we extend the intent prompt to support the broader label inventory.

\subsection{Intent Recognition}
\label{app:prompt-intent}

We use the following prompt for all SLURP-TN training and dev-test experiments reported in Tables~\ref{tab:grid} and~\ref{tab:synth}.

\begin{promptbox}{System prompt}
You are a spoken language understanding system for Tunisian Arabic (Tunisian
dialect; code-switching with French/English words is common).

Task: listen to the audio utterance and classify the speaker's INTENT. Choose
exactly one label from the fixed inventory below - do not invent new labels, do
not translate, do not explain.

Valid intent labels (23): Emails, addcontact, alarm\_query, alarm\_remove,
alarm\_set, email\_addcontact, email\_query, email\_querycontact,
email\_sendemail, general\_greet, general\_joke, general\_quirky, greet, joke,
news\_query, query, querycontact, quirky, sendemail, set, takeaway\_order,
takeaway\_query, weather\_query

OUTPUT FORMAT (valid single-line JSON, no markdown or extra text):
\{"intent": "<one label copied exactly from the list above>"\}
\end{promptbox}

\begin{promptbox}{User turn}
<audio>\\
Listen to the utterance and identify its intent.\\
Respond ONLY with a single-line JSON object: \{"intent": "<label>"\}
\end{promptbox}

\subsection{Slot Filling}
\label{app:prompt-slot}

We use the organizers' reference prompt for slot filling~\cite{elleuch2026slurptn}. This keeps the output format consistent across zero-shot and fine-tuned models and allows us to apply the same evaluation pipeline to all systems.

\begin{promptbox}{System prompt}
You are an automatic speech recognition and spoken language understanding
system for Tunisian Arabic (Tunisian dialect, written in Arabic script;
code-switching with French/English words is common).

Task: listen to the audio and output ONE line that is the exact spoken
transcription, with semantic slots marked inline using this scheme:\\
\hspace*{2em}<label> slot value >\\
A slot opens with its label in angle brackets and closes with a lone '>'. Words
outside any <label> ... > span are left as plain transcription.

Valid slot labels: <alarm\_type>, <app\_name>, <artist\_name>, <business>,
<business\_name>, <business\_type>, <date>, <device\_type>, <drink\_type>,
<email\_address>, <email\_folder>, <event\_name>, <food\_type>,
<general\_frequency>, <house\_place>, <ingredient>, <joke\_type>, <list\_name>,
<meal\_type>, <media\_type>, <movie\_name>, <news\_topic>, <order\_name>,
<order\_type>, <person>, <personal>, <personal\_info>, <place\_name>,
<relation>, <time>, <time\_zone>, <timeofday>, <transport\_type>,
<weather\_descriptor>

Output only the annotated transcription line: no translation, no explanation,
no surrounding quotes.
\end{promptbox}

\begin{promptbox}{User turn}
<audio>\\
Transcribe the audio with inline semantic slots as instructed.
\end{promptbox}

\subsection{Official Test Set}
\label{app:prompt-open}

Once the label-space mismatch of Appendix~\ref{app:open} was apparent, we also re-ran inference with an open inventory, to measure how much of the test set the model could place outside the six trained scenarios when permitted to. We modify only the prompt and keep the model unchanged, expanding the label inventory from the 23 released training labels to the full 60-label SLURP inventory and allowing \code{unknown} when the utterance does not match any available label. We also identify the six scenarios covered by the training data to discourage the model from mapping unseen intents to familiar in-domain labels. This run is diagnostic: our submitted predictions come from the 23-label prompt in Appendix~\ref{app:prompt-intent}, with the abstention rule in Appendix~\ref{app:open} applied on top.

\begin{promptbox}{System prompt}
You are a spoken language understanding system for Tunisian Arabic (Tunisian
dialect; code-switching with French/English words is common).

Task: listen to the audio utterance and classify the speaker's INTENT. Choose
exactly one label from the fixed inventory below - do not invent new labels, do
not translate, do not explain.

Valid intent labels (60): alarm\_query, alarm\_remove, alarm\_set,
audio\_volume\_down, audio\_volume\_mute, audio\_volume\_other,
audio\_volume\_up, calendar\_query, calendar\_remove, calendar\_set,
cooking\_query, cooking\_recipe, datetime\_convert, datetime\_query,
email\_addcontact, email\_query, email\_querycontact, email\_sendemail,
general\_greet, general\_joke, general\_quirky, iot\_cleaning, iot\_coffee,
iot\_hue\_lightchange, iot\_hue\_lightdim, iot\_hue\_lightoff, iot\_hue\_lighton,
iot\_hue\_lightup, iot\_wemo\_off, iot\_wemo\_on, lists\_createoradd,
lists\_query, lists\_remove, music\_dislikeness, music\_likeness, music\_query,
music\_settings, news\_query, play\_audiobook, play\_game, play\_music,
play\_podcasts, play\_radio, qa\_currency, qa\_definition, qa\_factoid,
qa\_maths, qa\_stock, recommendation\_events, recommendation\_locations,
recommendation\_movies, social\_post, social\_query, takeaway\_order,
takeaway\_query, transport\_query, transport\_taxi, transport\_ticket,
transport\_traffic, weather\_query

If the utterance does not fit ANY label above, answer exactly: unknown

IMPORTANT: this test set covers the FULL inventory above, which is much broader
than the six scenarios (alarm, email, general, news, takeaway, weather) you may
be most familiar with. Many utterances are about music, calendars, lists,
IoT/smart-home devices, transport, cooking, social media, general
question-answering or audio volume. Classify what you actually hear.\\
Do NOT use general\_quirky as a catch-all: reserve it for genuinely nonsensical
or unanswerable chit-chat. If an utterance has a clear topic that is not in the
list, answer unknown instead of general\_quirky.

OUTPUT FORMAT (valid single-line JSON, no markdown or extra text):
\{"intent": "<one label copied exactly from the list above, or unknown>"\}
\end{promptbox}

\end{document}